# A De-raining semantic segmentation network for real-time foreground segmentation

Fanyi Wang[1*], Yihui Zhang[2]

[1]Zhejiang University   [2]Henan University of Science and Technology

**Abstract**
Few researches have been proposed specifically for real-time semantic segmentation in rainy environments. However, the demand in this area is huge and it is challenging for lightweight networks. Therefore, this paper proposes a lightweight network which is specially designed for the foreground segmentation in rainy environments, named De-raining Semantic Segmentation Network (DRSNet). By analyzing the characteristics of raindrops, the MultiScaleSE Block is targetedly designed to encode the input image, it uses multi-scale dilated convolutions to increase the receptive field, and SE attention mechanism to learn the weights of each channels. In order to combine semantic information between different encoder and decoder layers, it is proposed to use Asymmetric Skip, that is, the higher semantic layer of encoder employs bilinear interpolation and the output passes through pointwise convolution, then added element-wise to the lower semantic layer of decoder. According to the control experiments, the performances of MultiScaleSE Block and Asymmetric Skip compared with SEResNet18 and Symmetric Skip respectively are improved to a certain degree on the Foreground Accuracy index. The parameters and the floating point of operations (FLOPs) of DRSNet is only 0.54M and 0.20GFLOPs separately. The state-of-the-art results and real-time performances are achieved on both the UESTC all-day Scenery add rain (UAS-add-rain) and the Baidu People Segmentation add rain (BPS-add-rain) benchmarks with the input sizes of 192*128, 384*256 and 768*512. The speed of DRSNet exceeds all the networks within 1GFLOPs, and Foreground Accuracy index is also the best among the similar magnitude networks on both benchmarks.

## 1. Introduction

Currently, semantic segmentation networks emerge in an endless stream and have been widely used in production and life. The current networks are mainly developing towards two directions of "being lighter and faster under the premise of certain performance" and "breaking through the current performance indicators". In recent years, with the advance of AI technology, the update and iteration of lightweight networks [1-13] have become more and more rapid. For example, ENet [1], CGNet [2], ContextNet [3], LEDNet [4], DFANet [5], FDDWNet [6] and so on, all committed to achieve a balance between accuracy and model complexity. The design of lightweight networks mainly has the following structural techniques: use asymmetric convolution, pointwise convolution, larger scale down-sampling, dropout, reduce channels, replace concat with add in channel merging, replace deconvolution with interpolation up-sampling, of course, an ingenious structure is the most important. In terms of breaking through performance index, use as large an input image as possible, use superior backbone, Lovaze-Softmax loss [14], learnable convolution methods such as DUpsampling [15] for up-sampling, multi-scale image and multi-model fusion for inference, set learning rate warmup method and optimize the inference results utilizing conditional Random Fields (CRF) [16] or Markov Random Fields (MRF) [17, 18] and so on, all can improve the performance of model to a certain degree. For different application environments, there exists well-recognized good-job networks. For example, the symmetric encoder-decoder structure of UNet [19] is effective for foreground segmentation with a simple background, combined atrous spatial pyramid pooling (ASPP) module with encoder-decoder structure, and based on the high-performance backbone [20-22], DeepLab v3+ [23] has an irreplaceable position in the multi-class segmentation with complex backgrounds. However, few attention has been paid to lightweight semantic segmentation network research in rainy environments, and the demand in this area is huge. For example, a major problem faced by autonomous driving and portrait segmentation is the foreground segmentation in rainy environments. The background of rainy environments is more complex to segment than sunny or indoor ones. The lightweight networks have limited parameters and certain structural design, most of them can only adapt to some special environments, and show poor segmentation performance in rainy environments. Aimed at this problem, we design and propose a lightweight semantic segmentation network named DRSNet, which is specifically for real-time foreground segmentation in rainy environments.

Although the use of tricks can improve the performance of the model, it can not be explained that the performance improvement is brought by the structural advantages. Therefore, in the experiments, we try to avoid the tricks as much as possible and speak with the structure of the network. In order to verify DRSNet's real-time segmentation performance on roads and portraits in rainy environments, the UAS dataset covering various weather and environmental conditions and Baidu's open source multi-scenes portrait

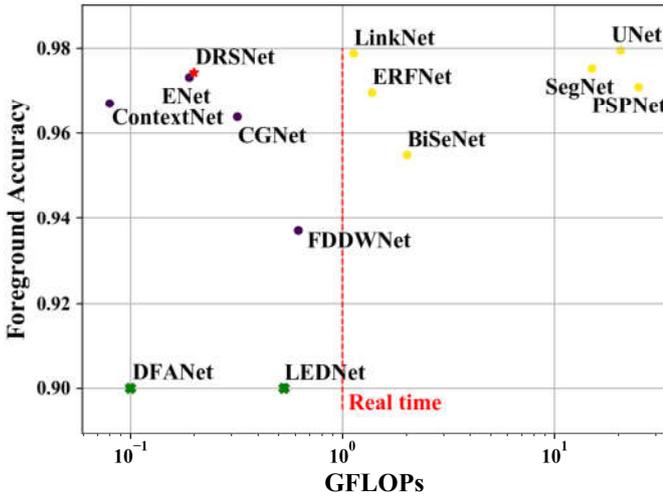
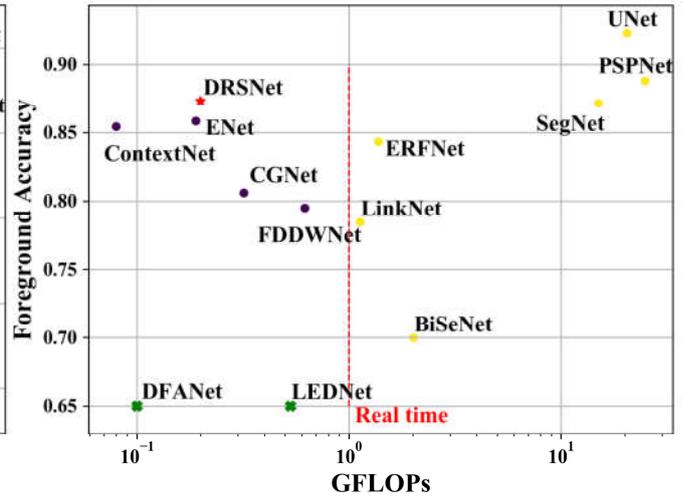

(a) Experiment results on UAS-add-rain benchmark.     (b) Experiment results on BPS-add-rain benchmark.

**Fig. 1** GFLOPs vs Foreground Accuracy performance on UAS-add-rain and BPS-add-rain benchmarks, **a** is the experiment results on UAS-add-rain benchmark and **b** is the experiment results on BPS-add-rain benchmark. In this paper, we define the real-time boundary as 1GFLOPs which is shown by the red dotted line. Purple circles represent lightweight networks, green forks stand for failed ones (we define the Forground Accuracy lower than 0.9 for UAS-add-rain and 0.65 for BPS-add-rain as failed ones) and our DRSNet is marked out with a red star.

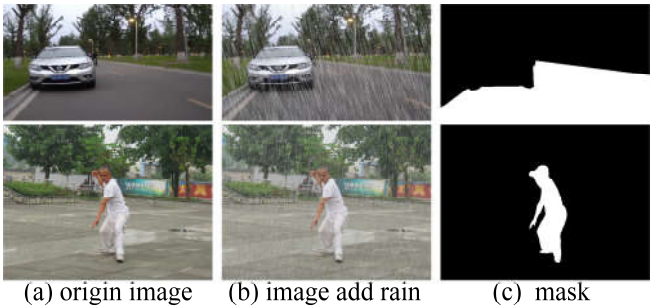

(a) origin image    (b) image add rain    (c) mask

**Fig. 2** Examples of UAS-add-rain benchmark (first row) and BPS-add-rain benchmark(second row).

dataset BPS are selected, artificial rain is added to UAS [24] and BPS [25] datasets to make up of our final benchmarks. A variety of current stat-of-the-art lightweight networks [1-6] and some relatively heavy networks [7, 12, 13, 19, 26, 27] are chosen as baselines for comparison so as to verify the effectiveness of our proposed DRSNet.

As shown in Fig. 1a and Fig. 1b, our DRSNet is marked out with a red star, and due to the calculation amount can more objectively reflect the real-time performance, we define the real-time boundary as 1GFLOPs which is shown by the red dotted line. Among all lightweight networks with calculation amount less than 1GFLOPs, the Foreground Accuracy index of DRSNet is the highest on UAS-add-rain benchmark. In addition, the light-weight networks [1-6] those express stat-of-the-art results on datasets such as cityscape [28] and camvid [29] do not adapt well to the rainy environments, and their segmentation performances are not as well as DRSNet. What's more, compared with the stat-of-the-art UNet, the calculation amount of DRSNet is less than 1%, but can achieve similar segmentation performance as UNet.

Illustrated by Fig. 1b, the overall Foreground Accuracy of the BPS-add-rain benchmark is lower than that of the UAS-add-rain benchmark, that is, BPS-add-rain benchmark is more challenging for segmentation. However, DRSNet still shows the best performance among all lightweight networks with less than 1GFLOPs. In order to reduce the amount of parameters and calculation, the lightweight network has the inevitable weakness of robustness. Compared with the networks with more than 1GFLOPs in Fig. 1, the segmentation effect of our DRSNet on both benchmarks can also reach the upper middle level. What's more, labeled by green fork in Fig. 1a and Fig. 1b, LEDNet and DFANet can not converge on both benchmarks (for comparison, the Foreground Accuracy is marked as 0.9 in Fig. 1a and 0.65 in Fig. 1b). A conclusion can be drawn that the semantic segmentation task with rainy environments has great challenge for lightweight networks, thus related research is of great necessity.

As shown in Fig. 2, the two selected benchmarks only have foreground and background scenes. The foreground of UAS-add-rain is the road, and the foreground of BPS-add-rain is the human being. Since the purpose of our segmentation task is to accurately extract the foreground, this article uses Foreground Accuracy as the index of segmentation performance. Considering that in inference procedure, different networks have different GPU occupancy rates, and the computing density of the networks are different which has been proved in experiments that sometimes the FPS of lightweight networks will be lower than that of heavyweight networks. Therefore, we use the calculation amount GFLOPs as the evaluation index of the real-time performance of the model.

Our contributions can be summarized as following 3 points:
1. Propose a new end-to-end lightweight semantic segmentation network named DRSNet, which shows a high-performance on foreground segmentation in rainy environments. By analyzing the shape and grayscale characteristics of raindrops in images, the MultiScaleSE Block is designed for encoding, multi-scale convolutions and dilated convolutions are used to increase semantic information and receptive field, SE attention operation is performed to learn the weights of each channels. And the effectiveness of MultiScaleSE Block has been proved through experiments. In order to make full use of encoder information and mix the semantics between different encoder and decoder layers, Asymmetric Skip is applied to encoder and decoder layers. And the effectiveness of Asymmetric Skip has been proved through experiments as well. What's more, the proposed DRSNet series(A, B and C) can adapt well to input images with different scales, such as 192*128, 384*256 and 768*512, and can achieve real-time performances on a single GTX1080Ti.
2. On the two add-rain benchmarks, compared to the networks with similar parameters and computational magnitude, our DRSNet shows the best performance on Foreground Accuracy index. To a certain degree, our DRSNet is lightweight but sophisticated.
3. We summarize some structural skills for designing of lightweight networks, as well as some training and structural skills for improving the performance of semantic segmentation networks.

This article will introduce the related work of the predecessors in section 2, including the traditional unsupervised segmentation methods and the learning-based methods emerging after 2015. In section 3, we introduce the implementation details of our DRSNet. Based on the shape and grayscale characteristics of raindrops in rainy images, we design and analyze each module structure in DRSNet in a targeted manner. The section 4 introduces the experimental parameter settings involved in this article and analyzes the experimental results, including the performance comparison experiments between DRSNet and other networks on the two benchmarks, the control experiments that prove the effectiveness of the MultiScaleSE block and the Asymmetric Skip structure, and of course, a discussion over the performance of our DRSNet. Eventually, section 5 summarizes the works of this article.

## 2. Related works

Due to the limitation of computer hardware, early segmentation tasks can only achieve simple segmentation on grayscale images. This stage was generally based on unsupervised learning, mainly by extracting the low-level features of the image for segmentation, and the segmentation results have no semantic annotation. And the representative algorithms have Ostu [30], FCM emerged [31], watershed [32], N-Cut [33] and so on. According to the segmentation methods, it can be divided into eight kinds which are listed as follows: based on threshold, region, region growth, region split and merge, edge detection, wavelet method, Markov Random Field model and genetic algorithm.

It was not until the groundbreaking work FCN [34] proposed in 2015 that for the first time a fully connected convolutional network was used for semantic segmentation of images, and then the semantic segmentation network[35-40] emerged like a mushroom. UNet, which adopts a symmetric skip structure, occupies the top position in the field of medical image segmentation, and SegNet, in which index pooling is applied to solve the problem of loss of information in the encoding process to a certain extent. Atrous convolution, CRF, ASPP, global average pooling those classic operations are proposed in DeepLab series [23, 41-43], and the latest DeepLab v3+ [23] has become a commonly used baseline in complex semantic segmentation events.

With the increasing migration trend of semantic segmentation network to mobile terminal, lightweight networks are gradually developing towards achieving a balance between accuracy and complexity. For example, ENet in 2016 only down-samples the input image four times, uses asymmetric convolution to reduce the amount of parameters, and shows that semantic segmentation is feasible on embedded devices. The main idea of CGNet in 2018 is deep and thin, it down-samples eight times, uses PReLU [44] activation function instead of ReLU, and the stat-of-the-art result is obtained on the cityscape benchmark. The DFANet proposed by Mgvii in 2019 is carried on the mobile terminal, it applies the Deep Feature Aggregation (DFA) structure to fully utilize the high-level feature information of the network, and the use of DFA lightweight feature aggregation structure not only reduces the amount of calculation by 7 times, but also breaks through real-time calculation boundary.

Although current lightweight semantic segmentation networks are frequently emerging, there are few studies on lightweight semantic segmentation networks specifically for rainy environments. Compared with the object detection field, some De-raining [45-47], De-Hazing [48-50] algorithms have been applied to optimize the detection results for a long time, and indeed have a certain effect. Compared with sunny or indoor environments, the difficulty of segmentation in rainy environments are undoubtedly increased. However, its application scenarios are wide, for example, one of the major problems encountered in the development of driverless driving is the road segmentation in rainy days, and portrait photography in rainy days may encounter inaccurate segmentation which will result in the failure of beauty algorithm. Due to the limited amount of parameters and calculation, the specific network structure, most of the lightweight networks can only have excellent segmentation performance for specific scenarios, which has been verified in our experiments, so researches for lightweight semantic segmentation in rainy environments is of great necessity.

## 3. Proposed network

The framework of proposed network DRSNet is illustrated in Fig. 3. It adopts the classical encoder-decoder structure in semantic segmentation, utilizes bilinear interpolation to downscale and upscale. As shown in Fig. 3, the number on the upper left corner of each module indicates the scaling ratio of the input feature map relates to the input image, and the number on the upper right corner of the module indicates the channel numbers of output feature map. Encoder is the process of modulating the semantic information of the image, each down-sampling operation will inevitably result in the loss of

information，and in our work, the information is supplemented by doubling the number of output channels. Decoder is the demodulation process of semantic information. In order to make full use of the semantic information in the encoder part and mix the semantics of different semantic layers to increase the amount of information in the decoder process, Asymmetric Skip is applied to encoder and decoder layers. Firstly, encoder's feature map is up-sampled through bilinear interpolation, then apply pointwise convolution to increase nonlinearity, finally, element-wise addition is performed with the corresponding lower-level decoder layer. And it is proved by control experiments that the asymmetric skip structure can more effectively integrate information between different semantic layers than Symmetric Skip in rainy environments with limited parameters.

Based on Fig. 3, the encoder and decoder procedure are introduced as follows. Firstly, The channels of input image is increased from 3 to 24 by conducting DoubleConv Block, and then performs 4 times bilinear interpolation down-sampling. After that, use MultiScaleSE Block to modulate and extract the semantic information, and each modulation performs 2 times bilinear interpolation down-sampling to obtain more advanced semantic information until the down-sampling scale reaches $1/16$ of the original size. After that, comes to decoder process, the $1/16$ scale feature map is interpolated 2 times at first, then the number of channels is reduced to half by PointwiseConv Block, after that, the semantic information is demodulated using DoubleConv Block, next, Asymmetric Skip is applied to merge the information of encoder and decoder layers. Loop the above operation until the feature map scale is restored to the original size, and then pass through NeckConv Block and PointwiseConv Block respectively. Ultimately, sigmoid activation function is utilized to predict the category of each pixel.

### 3.1 MultiscaleSE Block

Our DRSNet mainly has two innovative structures, MultiScaleSE Block and Asymmetric Skip. Asymmetric Skip is relatively simple and has been introduced before. Next, we will introduce and analyze MultiScaleSE Block in detail.

First of all, the shape and grayscale characteristics of the raindrops in the image are analyzed. Due to the camera delay effect, there will be motion blur along the motion trail of raindrops on the image, to be specific, there exists an elongated shadow along a certain direction. And the overall grayscale is bright and has certain transparency. For the solution of motion blur, given the assumption that an appropriate receptive field and utilize multi-scale convolutions will help. As the input image is merely 192*128 and the maximum width of simulated rain drops is only 7, therefore, $1\times1$, $3\times3$, $5\times5$ convolution kernels are used for experiments. According to [51], any $n\times n$ convolution can be replaced by a $n\times1$ convolution followed by a $1\times n$ convolution, which is called asymmetric convolution, and the computational cost decreases dramatically as $n$ grows. Therefore, our DRSNet uses asymmetric convolution instead of $n\times n$ convolution to reduce parameters and ensure performance meanwhile, and the dilation rate of convolution kernel is set to be 2. Aimed at the gray characteristics of raindrop images, the channel attention mechanism is recommended. From this, the form of MultiScaleSE Block is obtained. As shown in Fig. 4, the convolution results of different kernel scales are concated, this process fully encodes the semantic information, and then the concated result undergoes batch normalization, after that, the SE module updates the weights of channels so as to pay more attention to the more significant ones.

Experiments are carried out on three kinds of MultiScaleSE Blocks: A、B and C. Due to the semantic information obtained by convolutions of different scales is different, it is necessary to find out the most favorable part of feature maps for semantic segmentation, therefore we use the SE attention module to learn the weight of each channels of the feature map after concat, more favorable to the segmentation result, greater the weight. The latter half part of the three structures is a SE attention module. The structure difference lies in the choice of the number of convolution channels and the scale of the convolution kernels in the first half part.

The structure of MultiScaleSE Block A is illustrated in Fig. 4a, the channel numbers of input is reduced by $1\times1$ convolution firstly, then conducts $3\times1$ and $1\times3$ convolutions in order, after that, connects with the output of $1\times1$ convolution. Different from MultiScaleSE Block A, MultiScaleSE Block B divides two branches from the input directly, as can be seen in Fig. 4b, an extra 1*1 convolution operation is added to the right branch. Three branches are divided from the input in MultiScaleSE Block C, and the divided channels respectively pass through the $1\times1$, $3\times3$, $5\times5$ asymmetric convolutions and then connect together. It has been proved through experiments in section 4 that the structure shown in Fig. 4a can obtain the highest segmentation index than the other two structures shown in Fig. 4b and Fig. 4c with small input size, however, with the increase of input size, the advantage of MultiScaleSE Block C gains gradually. The DRSNet mentioned in this article defaults to DRSNet(A).

### 3.2 Doubleconv Block and Neckconv Block

As illustrated in Fig. 5, the structure of DoubleConv Block and NeckConv Block are almost the same, the only difference lies in the number of output channels after the first convolution, DoubleConv Block is $C_{out}/2$, NeckConv Block is $C_{in}/2$, both have the purpose of reducing the number of channels in the convolution process, both use asymmetric convolution replaces symmetric convolution. When $C_{out}/2$ and $C_{in}/2$ are equal, DoubleConv Block and NeckConv Block are identical. The DoubleConv Block is used once for the input image during the encoder process, the purpose is to expand the number of image channels so as to enrich the semantic information. The DoubleConv Block mainly plays the role of semantic information demodulator in the decoder process and the NeckConv Block is only used in the last link of the decode process.

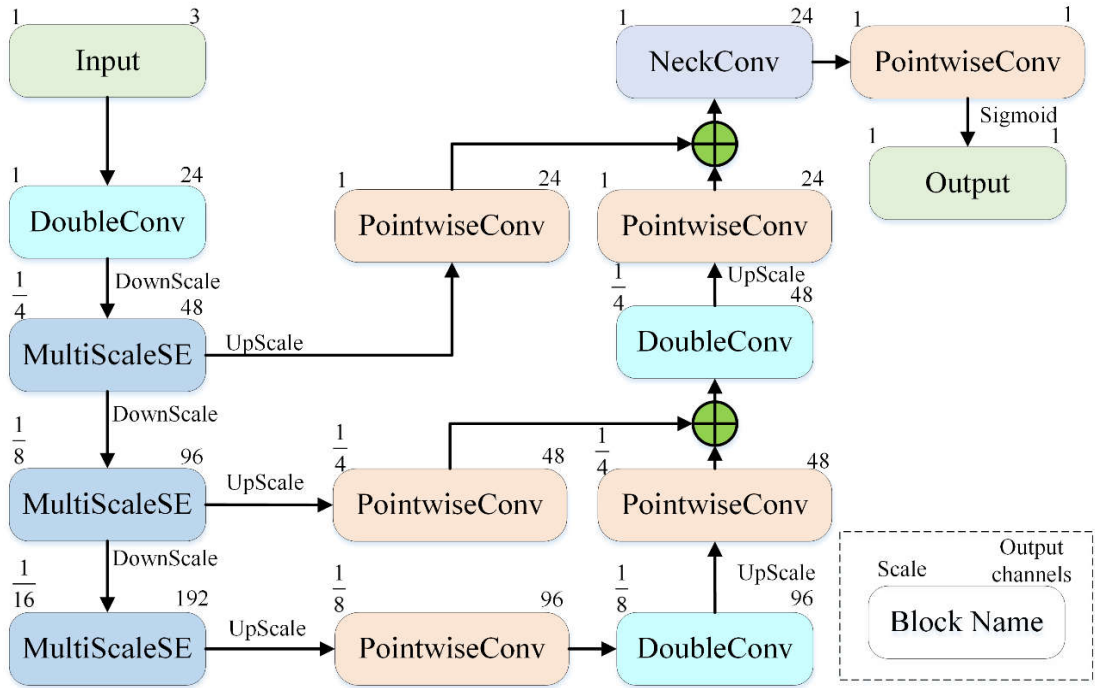

**Fig. 3** Flow chart of DRSNet, different colors represent different blocks, the output scale of the block is labeled on the upper left corner and the number of output channels of the block is labeled on the upper right corner.

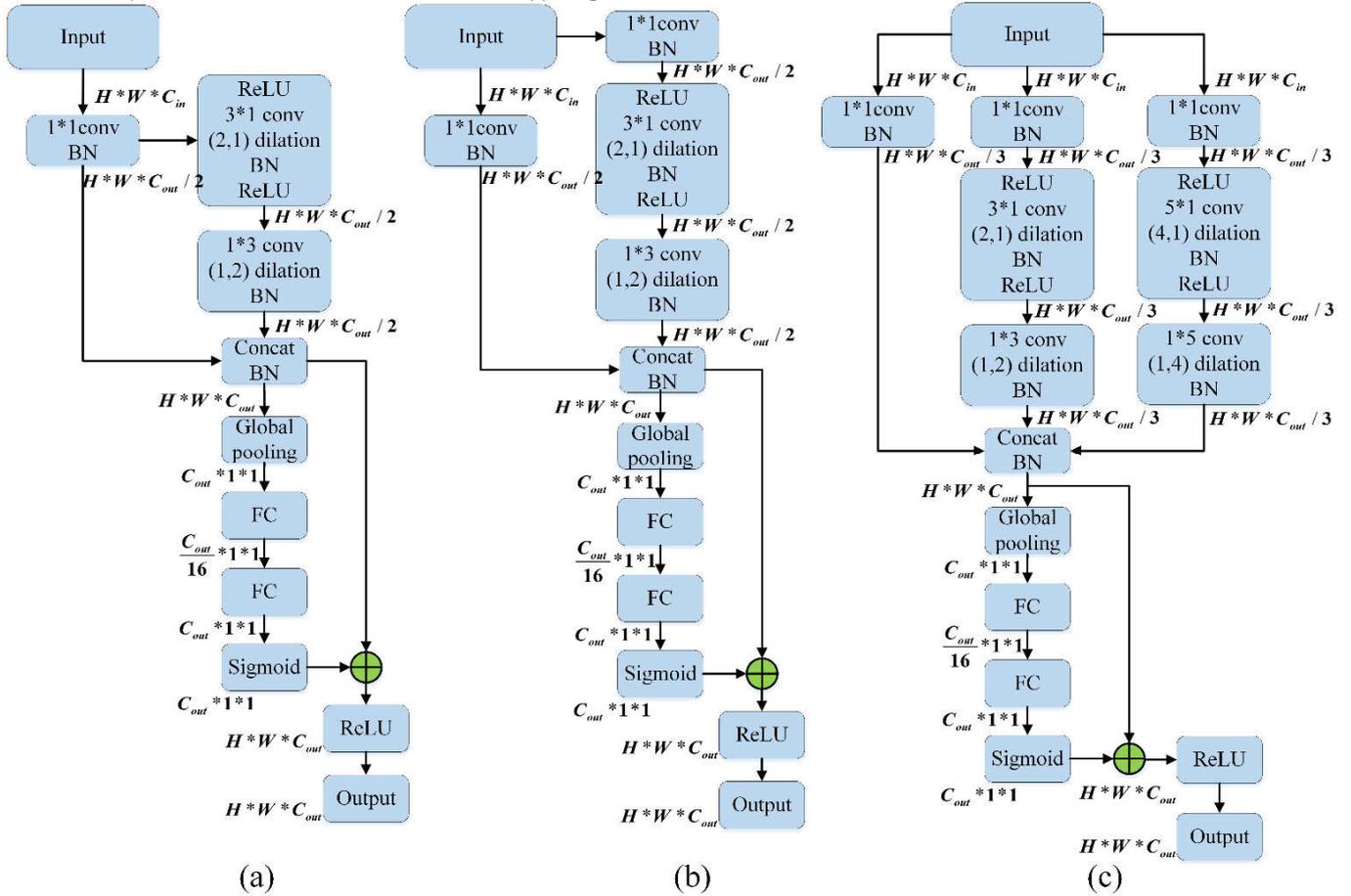

**Fig. 4** **a** is the structure of MultiScaleSE Block A, **b** is the structure of MultiScaleSE Block B and **c** is the structure of MultiScaleSE Block C.

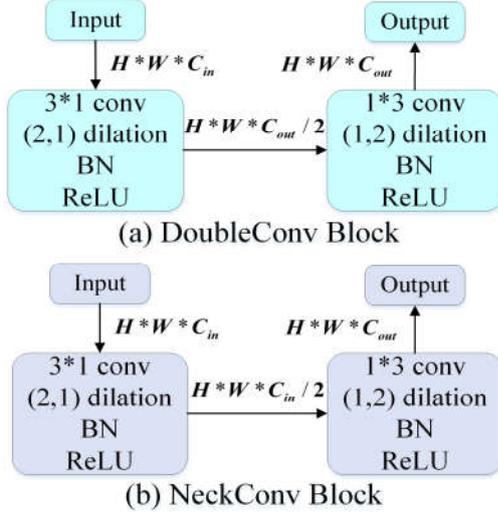

**Fig. 5 a** Structures of DoubleConv Block and **b** NeckConv Block.

## 4. Experiments

Generally, the lower the image resolution, the faster the prediction speed, however, most of the current lightweight networks conducted experiments on large image scales, for example, the input image scale used in the ENet experiment is 512*512, and in CGNet experiment is 680*680, or the structural is designed for large-scale images merely, such as LEDNet, the maximum down-sampling scale of which is up to 64 times, The consequence brought by this structure is that it is easy to cause the high-level semantic information to be too abstract to distinguish on small-scale images or objects. Nevertheless, our DRSNet can achieve good performance on 192*128 low-resolution input images. And for large-scale image segmentation, such as 384*256 and 768*512, stat-of-the-art results can be achieved as well. Further more, by superimposing MultiScaleSE Blocks and down-sampling operations, our DRSNet series can be qualified for even larger-scale images.

### 4.1 Experiment settings

Our experiments are based on the two datasets of UAS and BPS, and Gaussian blur filter is applied to the original image to simulate raindrop noise. Considering that due to the delay of the camera, raindrops have motion blur and are affected by the wind, therefore, a rectangular Gaussian kernel is used and a certain tilt angle is added to the Gaussian kernel. In addition, the raindrops are transparent, so the transparency parameter is added to generate raindrop noise. In order to verify the De-raining ability of DRSNet, heavy raindrop noise is added, and the parameters of Gaussian blur kernel used in simulation are shown in Table 1. In order to simulate the rainy environment more realistically [52], the raindrop parameters are randomly selected within a certain range, and the length is set to be an integer in the interval [40, 60]. The fuzzy kernel of raindrop noise is randomly selected from the set {3, 5, 7}, the raindrop slope angle interval is in the range of [-30°, 30°], the transparency interval is in the range of [0.6, 0.9], and the probability interval for raindrops arise on a certain pixel is among [1/250, 1/245]. Due to the length and width of raindrop noise, theoretically, even in the most sparsest case, raindrops can cover $40 \times 3 \times 1 \div 250 = 12/25$ of the entire image, therefore, we can simulate a heavy rain environment realistically. The hardware and software platforms configuration are shown in Table 2.

**UAS dataset:** The full name of the UAS dataset is UESTC All-Day Scenery, which provides all-weather road pictures and corresponding binary labels, and the foreground is roads. The entire data set contains four kinds of weather (dull weather, night weather, rainy weather and sunny weather), with a total of 6380 images, the original resolution is 360*640, most of the rainy weather images are taken after rain, so there is almost no raindrop noise in the background. In this article, artificial raining operation is added to the UAS to verify the performance of DRSNet for segmentation of roads in rainy environments.

**BPS dataset:** This data set is mainly used for the segmentation of human being, and it was derived from the Baidu browser, consists of 5387 pictures, mostly acquired outdoors in sunny weather. The original resolutions of BPS dataset are inconformity. In this article, the artificial raining operation is added to the BPS to verify the performance of DRSNet for segmentation of human being in rainy environments. Since the scales between the original images vary greatly, the tilt angle after resize will also changes greatly, which will increase the difficulty of semantic segmentation to a certain extent.

In order to prove that the improvement of segmentation performance is brought by the structure of DRSNet, this article only uses one training trick, learning rate warmup. In order to prove that DRSNet can achieve excellent segmentation effect on low-resolution images, the resolution of the input image was uniformly resized to 192*128 in the experiment, and the input size of 384*256 and 768*512 are experimented as well so as to prove that our DRSNet can adapt well to input images with different scales. Use SGD to optimize the gradient, and the momentum is set to be 0.9, weight decay is set to be 0.0001, initial learning rate is set to be 0.008, cosine annealing warmup is used, batch size is set to be 20, a total of 20 epochs are trained, and the ratio of training and testing data is 9:1, loss function selects BCEloss, and its expression is shown in formula (1).

$$BCE = \sum_i -y_i \log y_i^{'} - (1-y_i) \log(1-y_i^{'}) \quad (1)$$

Since the two datasets selected in this article only have foreground and background, the purpose of segmentation is to accurately segment the foreground, so the Foreground Accuracy is selected as the segmentation performance index which means the Pixel Accuracy(PA) of the foreground, And the calculation method is as follows:

$$Foreground\ Accuracy\ =\ TP/(FP+TP) \quad (2)$$

**TP** (True Positive): The prediction is true. The prediction is a positive class and the truth is also a positive class.

**FP** (False Positive): The prediction is false. The prediction is a positive class but the truth is a negative class.

**Table 1** Simulation raindrop parameters.

| Kernel length/Pixels | Kernel width/Pixels | Angle | Transparency | Intensity |
|---|---|---|---|---|
| [40, 60] | {3, 5, 7} | [-30°, 30°] | [0.6, 0.9] | [1/250, 1/245] |

**Table 2** Hardware and software configuration.

| CPU | RAM | GPU | OS | Environment |
|---|---|---|---|---|
| Intel Xeon CPU E5-2643 v3 @3.4GHz | 64GB | GTX 1080Ti | Ubuntu 18.04 | Pytorch1.1.0 |

**Table 3** Experiment results on UAS-add-rain benchmark for the input size of 192*128.

| Method | Foreground Accuracy | mIoU | Time(ms) | Speed(FPS) | Para(M) | FLOPs(G) |
|---|---|---|---|---|---|---|
| ENet(2016)[1] | 0.9731 | 0.9227 | 13.4 | 74.4 | **0.35** | 0.19 |
| CGNet(2018)[2] | 0.9638 | 0.9036 | 15.1 | 66.4 | 0.49 | 0.32 |
| LEDNet(2019)[4] | Nan | Nan | Nan | Nan | 0.92 | 0.53 |
| DFANet(2019)[5] | Nan | Nan | Nan | Nan | 2.09 | 0.10 |
| FDDWNet(2019)[6] | 0.9370 | 0.8853 | 17.6 | 56.7 | 0.81 | 0.62 |
| ContextNet(2018)[3] | 0.9670 | 0.8968 | 6.8 | 146.8 | 0.87 | **0.08** |
| BiSeNet(2018)[8] | 0.9549 | 0.9113 | 4.3 | 234.3 | 12.40 | 2.02 |
| SegNet(2015)[9] | 0.9753 | 0.9344 | 4.9 | 203.8 | 29.44 | 15.04 |
| PSPNet(2017)[10] | 0.9709 | 0.9406 | 17.6 | 56.9 | 65.57 | 24.94 |
| UNet(2015)[11] | **0.9795** | **0.9410** | **4.1** | **243.9** | 31.04 | 20.52 |
| LinkNet(2017)[12] | 0.9788 | 0.9257 | 5.2 | 192.8 | 11.53 | 1.13 |
| ERFNet(2018)[13] | 0.9697 | 0.9132 | 9.0 | 110.8 | 2.06 | 1.38 |
| **DRSNet(A)** | **0.9746** | **0.9255** | **6.4** | **155.9** | **0.54** | **0.20** |
| DRSNet(B) | 0.9635 | 0.9054 | 5.9 | 169.5 | 0.55 | 0.20 |
| DRSNet(C) | 0.9725 | 0.9246 | 6.8 | 147.1 | 0.33 | 0.20 |

**Table 4** Experiment results on BPS-add-rain benchmark for the input size of 192*128.

| Method | Foreground Accuracy | mIoU | Time(ms) | Speed(FPS) | Para(M) | FLOPs(G) |
|---|---|---|---|---|---|---|
| ENet(2016)[1] | 0.8583 | 0.8387 | 13.2 | 89.8 | **0.35** | 0.19 |
| CGNet(2018)[2] | 0.8060 | 0.7945 | 14.6 | 81.5 | 0.49 | 0.32 |
| LEDNet(2019)[4] | Nan | Nan | Nan | Nan | 0.92 | 0.53 |
| DFANet(2019)[5] | Nan | Nan | Nan | Nan | 2.09 | 0.10 |
| FDDWNet(2019)[6] | 0.7949 | 0.8030 | 23.7 | 42.3 | 0.81 | 0.62 |
| ContextNet(2018)[3] | 0.7844 | 0.7904 | 6.1 | 164.2 | 0.87 | **0.08** |
| BiSeNet(2018)[8] | 0.6996 | 0.7600 | 4.8 | 247.2 | 12.40 | 2.02 |
| SegNet(2015)[9] | 0.8715 | 0.8600 | 5.1 | 233.4 | 29.44 | 15.04 |
| PSPNet(2017)[10] | 0.8879 | 0.8307 | 22.1 | 45.3 | 65.57 | 24.94 |
| UNet(2015)[11] | **0.9227** | **0.8850** | **3.8** | **264.2** | 31.04 | 20.52 |
| LinkNet(2017)[12] | 0.8543 | 0.8479 | 4.8 | 208.6 | 11.53 | 1.13 |
| ERFNet(2018)[13] | 0.8432 | 0.8437 | 9.6 | 104.7 | 2.06 | 1.38 |
| **DRSNet(A)** | **0.8731** | **0.8291** | **6.0** | **166.7** | **0.54** | **0.20** |
| DRSNet(B) | 0.8424 | 0.8347 | 5.7 | 175.4 | 0.55 | 0.20 |
| DRSNet(C) | 0.8712 | 0.8344 | 6.5 | 153.8 | 0.33 | 0.20 |

**Table 5** Experiment results of different input size on DRSNet A, B and C, and the optimal values for each input size are marked in red bold.

| Input size | Method | UAS-add-rain | | | | BPS-add-rain | | | | Para(M) | FLOPs(G) |
|---|---|---|---|---|---|---|---|---|---|---|---|
| | | Foreground Accuracy | mIoU | Time (ms) | Speed (FPS) | Foreground Accuracy | mIoU | Time (ms) | Speed (FPS) | | |
| **192*128** | DRSNet(A) | **0.9743** | **0.9213** | 6.4 | 166.7 | **0.8731** | 0.8291 | 6.0 | 166.7 | 0.54 | **0.20** |
| | DRSNet(B) | 0.9635 | 0.9187 | **5.9** | **169.5** | 0.8424 | **0.8347** | **5.7** | **175.4** | 0.55 | 0.20 |
| | DRSNet(C) | 0.9652 | 0.9208 | 6.8 | 147.1 | 0.8712 | 0.8346 | 6.5 | 153.8 | **0.33** | 0.20 |
| **384*256** | DRSNet(A) | **0.9746** | **0.9255** | 8.3 | 120.0 | **0.8837** | 0.8310 | 8.5 | 117.9 | 0.54 | **0.78** |
| | DRSNet(B) | 0.9635 | 0.9054 | **8.1** | **123.2** | 0.8341 | **0.8385** | **7.8** | **128.4** | 0.55 | 0.79 |
| | DRSNet(C) | 0.9725 | 0.9246 | 9.4 | 106.4 | 0.8508 | 0.8418 | 9.1 | 109.6 | **0.33** | 0.81 |
| **768*512** | DRSNet(A) | 0.9726 | 0.9099 | 8.8 | 113.9 | 0.8337 | 0.8010 | 8.8 | 113.3 | 0.54 | **3.13** |
| | DRSNet(B) | 0.9713 | 0.9076 | **8.7** | **115.3** | 0.8626 | 0.8078 | **8.4** | **118.7** | 0.55 | 3.17 |
| | DRSNet(C) | **0.9776** | **0.9300** | 9.8 | 101.8 | **0.8689** | **0.8431** | 9.3 | 107.9 | **0.33** | 3.22 |

**Table 6** Control experiments to verify the function of MultiScaleSE Block and Asymmetric Skip structures with the input size of 192*128.

| Method | Foreground Accuracy(UAS-add-rain) | Foreground Accuracy(BPS-add-rain) | Para(M) | FLOPs(G) |
|---|---|---|---|---|
| **DRSNet(A)** | **0.9746** | **0.8731** | 0.54 | 0.20 |
| SEResNet18 | 0.9679 | 0.8685 | 1.68 | 1.02 |
| Symmetric Skip DRSNet(A) | 0.9652 | 0.8593 | 0.52 | 0.18 |
| ResNet18(BottleNeck) | 0.9461 | 0.8609 | 3.53 | 4.36 |

**FN** (False Negative): The prediction is false. The prediction is a negative class but the truth is a positive class.
**TN** (True Negative): The prediction is true. The prediction is a negative class and the truth is also a negative class.

In the mIoU calculation formula, $k = 1$:

$$mIoU = \frac{1}{k+1} \sum_{i=0}^{k} \frac{p_{ii}}{\sum_{j=0}^{k} p_{ij} + \sum_{j=0}^{k} p_{ji} - p_{ii}} \quad (3)$$

### 4.2 Performance verification experiments

In experiments, 12 semantic segmentation networks are selected for comparison with our DRSNet, including 6 lightweight networks with less than 1GFLOPs and 6 networks with more than 1GFLOPs. The experiments were conducted on the UAS-add-rain and BPS-add-rain benchmarks. The experimental results are shown in Table 3 and Table 4 in which the input size are 192*128. The optimal results of each item are marked in red bold. The results of our proposed DRSNet are shown in black bold at the end of the Table 3 and Table 4. And in order to prove the universality of our DRSNet series towards different input scales, experiments were conducted on the input size of 192*128, 384*256 and 768*512, the results are illustrated in Table 5.

### 4.2.1 Experiments on UAS-add-rain

It can be seen from Table 3 that on the UAS-add-rain benchmark, UNet shows the highest Foreground Accuracy, reaches 0.9795, which is only a little bit higher than the 0.9746 of our DRSNet(A), but the parameter amount is almost 60 times of our DRSNet(A). It can be drawn from the results of FPS that some networks even with larger calculations are faster than the lighter ones. By monitoring GPU usage efficiency and memory consumption, it was found that when calculating FPS, networks with more parameters and larger calculations usually occupied more GPU computing resources and memory. Therefore, it is possible to obtain faster inference speed. From the perspective of practical application, the FPS parameter of the network is unreliable, and the FLOPs parameter is more favorable. The smallest FLOPs in Table 3 is owned by ContextNet proposed in 2018, whose calculation amount is only 0.08GFLOPs, which is 0.4 times of DRSNet(A), but in terms of Foreground Accuracy, it is 0.76 percentage points lower than our DRSNet. At the same time, on the UAS-add-rain benchmark, the prediction speed of DRSNet(A) is as high as 155.9 FPS which surpasses the other six lightweight networks within 1 GFLOPs. DRSNet(A) has less parameters and higher Foreground Accuracy than DRSNet(B). Although DRSNet(C) has fewer parameters, Foreground Accuracy performance is 0.21 percentage points lower than DRSNet(A) for the input size of 192*128, so as mentioned in section 3, we choose MultiScaleSE Block A as the default backbone. Compared with ENet, DRSNet series all have advantages in both Foreground Accuracy and parameter quantity, which further proves that the network structure proposed in this article has excellent performance for foreground segmentation in rainy environments.

Seven kinds of networks: ENet, CGNet, FDDWNet, LEDNet, BiSeNet, SegNet and our DRSNet are selected based on the experimental results shown in Table 3 and Table 4 to compare the visual effects of segmentation results. Based on the UAS-add-rain benchmark, as shown in Fig. 6, the first two input images are the daytime scenes and the last three ones are the night scenes, as can been seen, good performance can be achieved no matter by day or night in experiments. The segmentation results of LEDNet are not good and have obvious aliasing which has a certain relationship with its structure as has been mentioned before, the minimum down-sampling multiple of LEDNet is as high as 64 times, and it can not adapt well to the segmentation of small-scale input images. Among the light-weight networks, the performance of ENet is fine, and the overall segmentation effect of DRSNet(A) is better than ENet, and even can be better by using post-processing algorithms such as CRF for optimization. Compared with the other six networks in Fig 6, the foreground segmentation results of DRSNet(A) are visually superior to a certain degree.

### 4.2.2 Experiments on BPS-add-rain

Compared with the UAS-add-rain benchmark, the BPS-add-rain benchmark has less data, and the background is more complicated, so the foreground segmentation is more challenging. LEDNet and DFANet can not converge on this dataset. As can be seen from Table 4, for this dataset, UNet has the highest Foreground Accuracy, reaches 0.9227, but the parameter amount is also as high as 20.52GFLOPs. Although the Foreground Accuracy index of UNet is 4.96 percentage points higher than our DRSNet(A), the calculation amount is more than 100 times as well. UNet can't even run on the mobile terminal with limited computing resources, but our DRSNet proposed in this article can be competent. Compared with ENet, in Foreground Accuracy aspect, out DRSNet(A) is 1.48 percentage points higher than it. The prediction speed of DRSNet(A) on the BPS-add-rain benchmark is as high as 166.7 FPS. And DRSNet(A) has a higher Foreground Accuracy than DRSNet(B) and DRSNet(C) for the input size of 192*128.

Fig. 7 shows the segmentation results on the BPS-add-rain benchmark. This benchmark is more difficult than the UAS-add-rain for foreground segmentation, one is that the data volume is less, and the other is that the background is more complex and changeable. On this benchmark, the lightweight networks LEDNet and DFANet also failed, the segmentation results of them are poor, SegNet performs well but its calculation amount is as high as 15.04GFLOPs. Compared with DRSNet(A), although the calculation amount is only 1.3 percent of SegNet, the Foreground Accuracy index is even higher. Compared with the ENet of the similar magnitude, DRSNet(A) performs better in detail segmentation. These results are reliable enough to prove that DRSNet(A) has excellent performance for semantic segmentation in rainy environments for small input size.

## 4.3 Experiments for different input scales

As can been drawn from Table 6 that the proposed DRSNet series achieve stat-of-the-art results and possess strong robustness to images with different scales. And from Table 6, it can be found that with the increase of input scale, the segmentation effect of DRSNet(A) is gradually weakening, while the segmentation effect of DRSNet(C) is gradually increasing, which confirms that the previous assumption in 3.1 that appropriate receptive field of MultiScaleSE Block can improve the segmentation effect in rainy environments. And when it comes to the input size of 768*512, the Foreground Accuracy result of DRSNet (C) surpasses DRSNet (A) by 0.5 percentage points.

For UAS-add-rain and BPS-add-rain benchmarks，when the input size is 768*512, the frame rate of DRSNet (C) reaches 101.8FPS and 107.9FPS on a single GTX1080Ti respectively, which means that it is suitable for scenes such as real-time video segmentation. The Foreground Accuracy index are as high as 0.9776 and 0.8689 separately, for a lightweight network with only 0.33M parameters, the performance is excellent. What's more, this article also explores the temporal performance on 4K(3840*2160) images, and the frame rate can reach 15.6FPS and 16.2FPS for UAS-add-rain and BPS-add-rain benchmarks on a single GTX1080Ti respectively by utilizing DRSNet(B).

## 4.4 Structure validity verification experiments

The DRSNet series proposed in this article mainly relies on two structures: MultiScaleSE Block and Asymmetric Skip. In order to prove that these two structures are really helpful to improve the Foreground Accuracy index, three sets of control experiments were conducted and the results are shown in Table 6. Firstly, in order to prove that MultiScaleSE Block as an encoder is unique to the semantic segmentation for rainy environments, SEResNet18 is used as encoder for control experiment. The same hyper parameters are used during training process. For fairness, the number of SEResNet18 channels is reduced to same as MultiScaleSE Block, and the Asymmetric Skip structure is also used. The experiment results show that, on both the UAS-add-rain and BPS-add-rain benchmarks, the Foreground Accuracy of DRSNet(A) is higher than SEResNet18, and the amount of parameters and calculations are less as well.

In order to prove that Asymmetric Skip is indeed more helpful for the demodulation of semantic information in the decoder process, symmetric skip DRSNet(A) is used as a control group with the same training hyper parameters. As can be seen from Table 6, DRSNet(A) achieves better Foreground Accuracy than Symmetric Skip DRSNet(A) on the two benchmarks.

In order to make the ablation experiment sufficient enough to prove the excellent performance of the DRSNet series proposed in this article, the experiment of ResNet18 with BottleNeck is conducted and the results are shown in Table 6 that the performance of proposed DRSNet series are better than ResNet18 with BottleNeck.

Through the above three sets of control experiments, the conclusion can be drawn that the two structures: MultiScaleSE Block and Asymmetric Skip really play significant roles in the segmentation in rainy environments.

## 4.5 Discussion

It can be concluded from the above experimental results that the DRSNet series proposed in this article can achieve excellent segmentation performance on the two artificial add rain datasets, UAS-add-rain and BPS-add-rain. Compared with other lightweight segmentation networks, based on the assumption that proper receptive fields can improve the performance of foreground segmentation in rainy environments, we propose the MultiScaleSE Block based on the shape and gray-scale characteristics of raindrops in the image. And the assumption has been confirmed in Section 4.3 that with the increase of input size, DRSNet(C) with bigger receptive field shows advantage over DRSNet(A) gradually. Due to the tiny amount of parameters and calculations of DRSNet series, even for a large-scale input image of 768*512, they all can exceed 100FPS on a single GTX1080Ti, which means they are also applicable in the field of video segmentation. In addition, in order to verify the superiority of the two structures proposed in this paper, control experiments were carried out with SEResNet18, Symmetric Skip DRSNet (A) and ResNet18 with BottleNeck respectively. The experimental results illustrate that the two proposed structures in this paper do play an important role in foreground segmentation in rainy environments.

## 5 Conclutions

From the experiments, it was found that some lightweight segmentation networks failed to segment images with rainy environments, the specific performance is that the loss does not decrease or the final loss after convergence is still large. And the networks with larger parameters and calculation amounts are relatively more robust, they have better performance and can converge well on the benchmarks with rainy environments. This phenomenon further illustrates that it is necessary to design a targeted lightweight semantic segmentation network for rainy environments. Aiming at this problem, based on the well-designed structures, MultiScaleSE Block and Asymmetric Skip, a lightweight segmentation network named DRSNet is proposed for real-time foreground segmentation in rainy environments. Compared with the semantic segmentation networks within 1GFLOPs, on the UAS-add-rain and BPS-add-rain benchmarks, both the Foreground Accuracy and the inference speed of our DRSNet are the optimal, and the parameter and calculation amounts are only 0.54M and 0.2GFLOPs separately. On the UAS-add-rain benchmark, it's Foreground Accuracy reaches 0.9743, compared with UNet, which has more than 100 times the calculation amount of DRSNet, but the segmentation performance is similar. Under the Ubuntu 18.04 operation system, the inference speed reaches 155.9 FPS on the GTX 1080Ti GPU with the input size of 192*128. On the

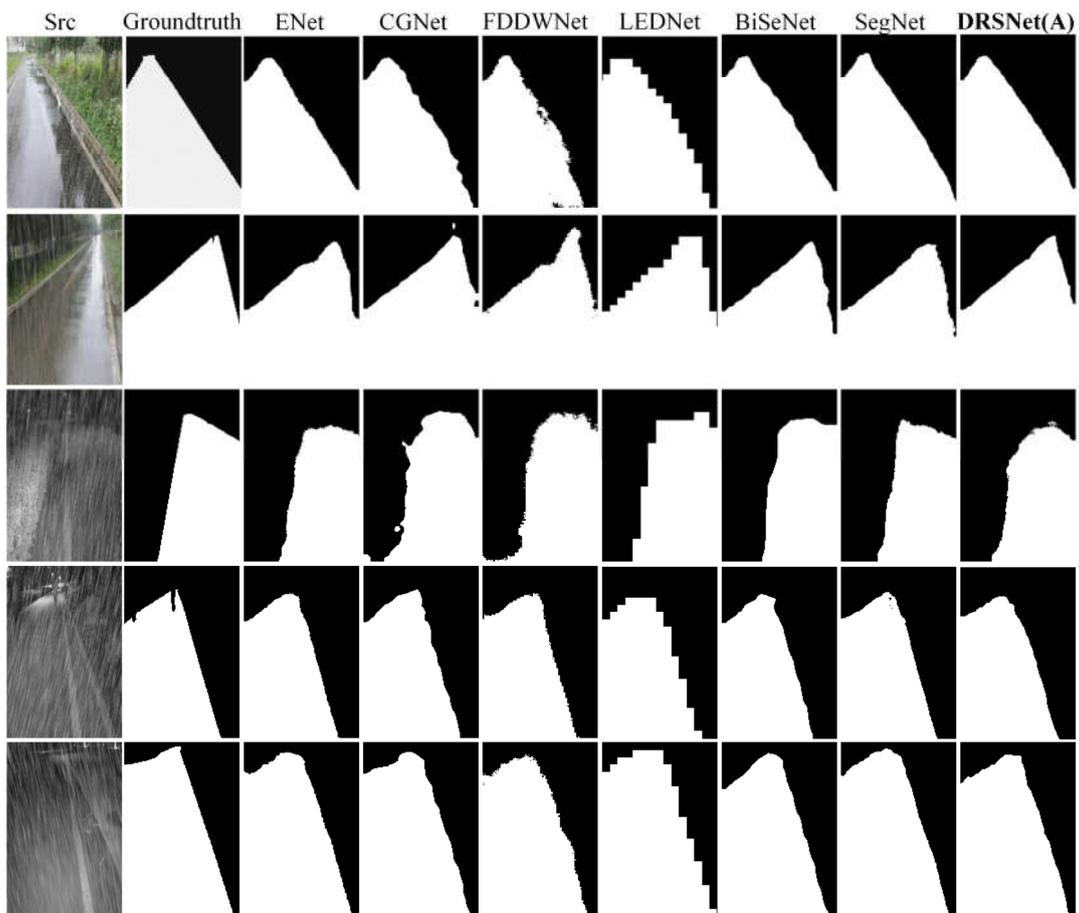
**Fig. 6** Comparison of the segmentation results between 7 kinds of networks on UAS-add-rain benchmark.

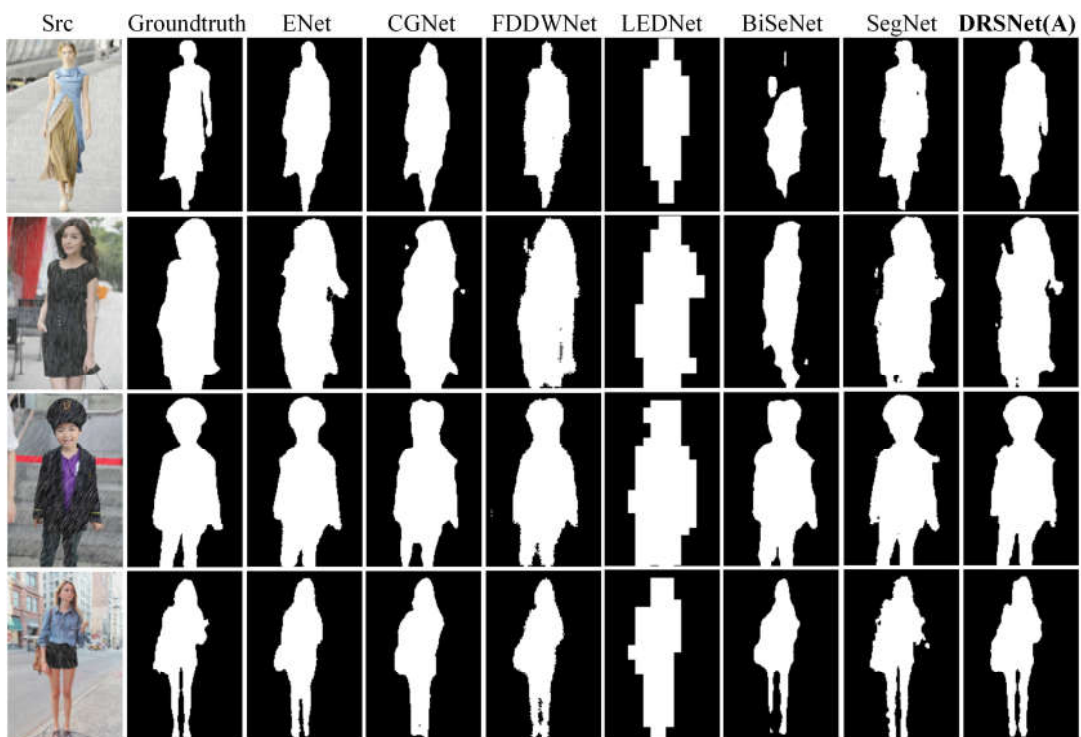
**Fig. 7** Comparison of the segmentation results between 7 kinds of networks on BPS-add-rain benchmark.

BPS-add-rain benchmark, Foreground Accuracy of DRSNet reaches 0.8731, and the inference speed is as high as 166.7 FPS. What's more, for input images with 384*256 and 768*512, the DRSNet series proposed in this article can also achieve stat-of-the-art result on both segmentation performance and real-time performance.

**Acknowledgements** The authors would like to thank the Associate Editor and the Reviewers for their constructive comments.

**Availability of code and data** The source code is released at https://github.com/dandingbudanding/DRSNet